\documentclass{article}

\PassOptionsToPackage{numbers}{natbib}
\usepackage[final]{neurips_bdl2019}

\def\compilefigures

\usepackage[utf8]{inputenc} 
\usepackage[T1]{fontenc}    
\usepackage{hyperref}       
\usepackage{url}            
\usepackage{booktabs}       
\usepackage{amsfonts}       
\usepackage{nicefrac}       
\usepackage{microtype}      
\usepackage{amsmath}
\usepackage{lipsum}
\usepackage{url}
\usepackage{subcaption}
\usepackage[subfigure]{tocloft}
\usepackage{pgfplots}
\usepackage[english]{babel}
\usepackage[linesnumbered,ruled,noend]{algorithm2e}
\usepackage{natbib}
\usepackage{booktabs}
\usepackage{colortbl}
\usepackage[english]{babel} 
\usepackage[]{hyperref}
\usepackage{bm}             
\usepackage{url}
\usepackage{placeins}
\usepackage[linesnumbered,ruled,noend]{algorithm2e}
\usepackage[font=small,labelfont=bf,tableposition=top]{caption}
\usepackage[bottom]{footmisc}
\usepackage{amsthm}
\usepackage{lscape}
\usepackage{xargs}          
\usepackage{siunitx}
\usepackage{textcomp}
\usepackage{catchfile}
\usepackage{lipsum} 
\usepackage{blindtext}

\newlength\figureheight
\newlength\figurewidth

\addto\extrasenglish{%

}

\DeclareCaptionLabelFormat{andtable}{#1~#2  \&  \tablename~\thetable}
\usepgfplotslibrary{groupplots}
\usetikzlibrary{arrows,shapes,automata,backgrounds,petri,positioning}
\usetikzlibrary{decorations.pathmorphing}
\usetikzlibrary{decorations.shapes}
\usetikzlibrary{decorations.text}
\usetikzlibrary{decorations.fractals}
\usetikzlibrary{decorations.footprints}
\usetikzlibrary{shadows}
\usetikzlibrary{calc}
\usetikzlibrary{spy}

\pgfplotsset{compat=newest}

\title{Efficient Approximate Inference with Walsh-Hadamard Variational Inference}

\author{%
Simone Rossi$^*$\\
\small
Department of Data Science\\
\small
EURECOM\\
\And
S\'ebastien Marmin$^*$ \\
\small
Department of Data Science\\
\small
EURECOM\\
\texttt{\small \{name.surname\}@eurecom.fr} \\
\And
Maurizio Filippone \\
\small
Department of Data Science\\
\small
EURECOM\\
}
\definecolor{mygreen}{rgb}{0.2, 0.7, 0.2}
\definecolor{myorange}{rgb}{0.9, 0.5, 0.0}

\definecolor{blue}{HTML}{5DA5DA}
\definecolor{orange}{HTML}{FAA43A} 
\definecolor{green}{HTML}{60BD68} 
\definecolor{pink}{HTML}{F17CB0} 
\definecolor{brown}{HTML}{B2912F} 
\definecolor{purple}{HTML}{B276B2} 
\definecolor{yellow}{HTML}{DECF3F} 
\definecolor{red}{HTML}{F15854} 
\definecolor{gray}{HTML}{4D4D4D}

\def\N{{\mathcal{N}}}

\newcommand{\R}{\mathbb{R}}

\newcommand{\T}{\top}

\newcommand{\diag}{\mathrm{diag}}

\newcommand{\norm}{\mathcal{N}}

\newcommand{\fvect}{\mathbf{f}}
\newcommand{\gvect}{\mathbf{g}}

\newcommand{\hvect}{\mathbf{h}}

\newcommand{\mvect}{\mathbf{m}}

\newcommand{\uvect}{\mathbf{u}}

\newcommand{\xvect}{\mathbf{x}}

\newcommand{\yvect}{\mathbf{y}}

\newcommand{\zerovect}{\mathbf{0}}

\newcommand{\phivect}{\boldsymbol{\phi}}

\newcommand{\muvect}{{\boldsymbol{\mu}}}

\newcommand{\sigmavect}{\boldsymbol{\sigma}}
\def\Sigmavect{{\bm{\Sigma}}}

\newcommand{\varepsilonvect}{\bm{\epsilon}}

\def\vect{{\mathrm{vect}}}
\newcommand{\bigO}{\mathcal{O}}

\newcommand{\name}[1]{{\textsc{#1}}\xspace}

\newcommand{\vi}{\name{vi}}

\def\fastfood{\name{fastfood}}

\newcommand{\gp}{\name{gp}}
\newcommand{\gps}{\textsc{gp}s\xspace}

\newcommand{\dnns}{\textsc{dnn}s\xspace}

\newcommand{\mcd}{\name{mcd}}

\newcommand{\relu}{{\textsc{r}}e\name{lu}}

\newcommand{\mnll}{\name{mnll}}

\newcommand{\rmse}{\name{rmse}}

\def\hvi{\name{whvi}}

\newcommand{\Exp}{\mathbb{E}}

\newcommand{\KL}{\name{kl}}

\DeclareMathAlphabet{\mathsfit}{\encodingdefault}{\sfdefault}{m}{sl}
\newcommand{\tens}[1]{\mathbf{{#1}}}

\def\Wtens{{\tens{W}}}

\def\Amatr{{\bm{A}}}
\def\Bmatr{{\bm{B}}}

\def\Gmatr{{\bm{G}}}
\def\Hmatr{{\bm{H}}}
\def\Imatr{{\bm{I}}}

\def\Smatr{{\bm{S}}}

\def\Wmatr{{\bm{W}}}
\def\Xmatr{{\bm{X}}}
\def\Ymatr{{\bm{Y}}}

\def\Pimatr{{\bm{\Pi}}}

\def\Sigmamatr{{\bm{\Sigma}}}
\def\Omegamatr{{\bm{\Omega}}}

\begin{document}

\makeatletter
\newcommand{\loadtikz}[1]{
    \filename@parse{#1}
    \let\fpath\filename@area
    \ifdefined\compilefigures
        \newcommand{\sometext}{\input{\filename@area\filename@base}}
        \sometext
    \else
        \includegraphics[]{\fpath\filename@base.pdf.pdf}
    \fi
}
\makeatother

\newcommand\blfootnote[1]{%
    \begingroup
    \renewcommand\thefootnote{}\footnote{#1}%
    \addtocounter{footnote}{-1}%
    \endgroup
}
\maketitle

\begin{abstract}
    Variational inference offers scalable and flexible tools to tackle intractable Bayesian inference of modern statistical models like Bayesian neural networks and Gaussian processes. 
For largely over-parameterized models, however, the over-regularization property of the variational objective makes the application of variational inference challenging. 
Inspired by the literature on kernel methods, and in particular on structured approximations of distributions of random matrices, this paper proposes Walsh-Hadamard Variational Inference, which uses Walsh-Hadamard-based factorization strategies to reduce model parameterization, accelerate computations, and increase the expressiveness of the approximate posterior beyond fully factorized ones.%
%

    \blfootnote{$^*$ Equal contribution}
\end{abstract}

\section{Introduction and Motivation}
\label{sec:introduction}


Scalable Bayesian inference for non-trivial statistical models is achieved with Variational Inference (\vi, \citet{Jordan99}). Variational Inference has continuously gained popularity as a flexible approximate inference scheme for a variety of models for which exact Bayesian inference is intractable.
Bayesian neural networks \citep{Mackay94,Neal1997} and in particular Deep Gaussian Processes with random features expansions \citep{Cutajar17,Damianou13} represent good examples of models for which inference is intractable, and for which \vi\xspace -- and approximate inference in general -- is challenging due to the nontrivial form of the posterior distribution and the large dimensionality of the parameter space \citep{Graves11,Gal16}.
Recent advances in \vi allow to effectively deal with these issues in various ways. 
A flexible class of posterior approximations can be constructed using, e.g., normalizing flows \citep{Rezende2015}, whereas large parameter space have pushed the research in the direction of Bayesian compression \citep{Louizos2017a,Molchanov2017}.


Let's consider a classic supervised learning task with $N$ input vectors and corresponding labels collected in $\Xmatr = \{\xvect_1, \ldots, \xvect_N \}$ and $\Ymatr = \{\yvect_1, \ldots, \yvect_N \}$, respectively; furthermore, let's consider \dnns with weight matrices $\Wtens = \left\{\Wmatr^{(1)}, \ldots, \Wmatr^{(L)} \right\}$, likelihood $p(\Ymatr | \Xmatr, \Wtens)$, and prior $p(\Wtens)$.
In the variational setting, a lower bound to the log-marginal likelihood can be derived as follows:
\begin{equation}
\label{eq:lowerbound}
\log\left[p(\Ymatr | \Xmatr)\right] \geq 
\Exp_{q(\Wtens)}[ \log p(\Ymatr | \Xmatr, \Wtens)  ]
- \KL\{q(\Wtens) \| p(\Wtens)\}\,,
\end{equation}
where $q(\Wtens)$ is a parameterized approximation  of the true posterior $p(\Wtens | \Xmatr, \Ymatr)$.
This bound has two undesirable charateristics: the first term, which acts as a model fitting term, depends on the choice of the form of the variational distribution. Simple distributions might not have enough expressiveness to efficiently characterize the posterior over model parameters. 
On the other hand, the latter term -- which acts as a regularizer -- penalizes solutions where the posterior is far away from the prior. 
This term is the dominant one in the objective in case of over-parameterized models, and as a result, the optimization focuses on keeping the approximate posterior close to the prior, disregarding the data fit term \citep{Bowman2016, Sonderby2016, Rossi2018}.


In this paper, we will try to solve both problems at once. 
Our proposal is to reparameterize the variational posterior over model parameters by means of a structured decomposition based on random matrix theory \citep{Tropp2011, Le13, Yu2016}. 
Without loss of generality, consider Bayesian \dnns with weight matrices $\Wmatr^{(l)}$ of size $D \times D$.
Compared with mean field \vi, our proposal has a number of attractive properties.
The number of parameters is reduced from $\bigO(D^2)$ to $\bigO(D)$, reducing the over-regularization effect of the \KL term in the variational objective.
We derive expressions for the reparameterization and the local reparameterization tricks, showing that, the computational complexity is reduced from $\bigO(D^2)$ to $\bigO(D \log{D})$.
Finally, unlike mean field \vi, we induce a (non-factorized) matrix-variate distribution to approximate the posterior over the weights, increasing flexibility by modeling correlations between the weights at a log-linear cost in $D$ instead of linear. 
The key operation within our proposal is the Walsh-Hadamard transform, and this is why we name our proposal Walsh-Hadamard Variational Inference (\hvi).

\paragraph*{Related Work.}
\hvi is inspired by a line of works that developed from random feature expansions for kernel machines \citep{Rahimi08}, which we briefly review here.
A positive-definite kernel function $\kappa(\xvect_i, \xvect_j)$ induces a mapping $\phivect(\xvect)$, which can be infinite dimensional depending on the choice of $\kappa(\cdot,\cdot)$.
Among the large literature of scalable kernel machines, random feature expansion techniques aim at constructing a finite approximation to $\phivect(\cdot)$.
For many kernel functions \citep{Rahimi08, Cho09}, this approximation is built by applying a nonlinear transformation to a random projection $\Xmatr\Omegamatr$, where $\Omegamatr$ has entries $\N(\omega_{ij}|0,1)$.
If the matrix of training points $\Xmatr$ is $N \times D$ and we are aiming to construct $D$ random features, that is $\Omegamatr$ is $D \times D$, this requires $N$ times $\bigO(D^2)$ time, which can be prohibitive when $D$ is large.
%
\fastfood \citep{Le13} tackles the issue of large dimensional problems by replacing the matrix $\Omegamatr$ with a random matrix for which the space complexity is reduced from $\bigO(D^2)$ to $\bigO(D)$ and time complexity of performing products with input vectors is reduced from $\bigO(D^2)$ to $\bigO(D \log D)$.
In \fastfood, the matrix $\Omegamatr$ is replaced by 
$
    \Omegamatr \approx \Smatr\Hmatr\Gmatr\Pimatr \Hmatr\Bmatr\,,
$    
where $\Pimatr$ is a permutation matrix, $\Hmatr$ is the Walsh-Hadamard matrix, whereas $\Gmatr$ and $\Bmatr$ are diagonal random matrices with standard Normal distributions and Rademacher ($\{\pm 1\}$), respectively.
%
%
%
%
%
$\Smatr$ is also diagonal with i.i.d. entries, and it is chosen such that
the elements of $\Omegamatr$ obtained by this series of operations are approximately independent and follow a standard Normal (see \citet{Tropp2011} for more details).
The Walsh-Hadamard matrix is defined recursively starting from
$\tiny
    H_2 = \begin{bmatrix}
        1 & 1  \\
        1 & -1 \\
    \end{bmatrix}
$ and then
$\tiny
    H_{2D} = \begin{bmatrix}
        H_D & H_D  \\
        H_D & -H_D \\
    \end{bmatrix}
$, possibly scaled by $D^{-1/2}$ to make it orthonormal.
The product of $\Hmatr\xvect$ can be computed in $\bigO(D\log D)$ time and $\bigO(1)$ of extra space using the in-place version of the Fast Walsh-Hadamard Transform \citep{Fino1976}.
\fastfood inspired a series of other works on kernel approximations
\citep{Yu16,Bojarski2017}, whereby Gaussian random matrices are approximated by a series of products between diagonal Rademacher and Walsh-Hadamard matrices, for example $\Omegamatr \approx \Hmatr \Smatr_1 \Hmatr \Smatr_2 \Hmatr \Smatr_3$.

\section{Walsh-Hadamard Variational Inference}
\label{sec:whvi}


\hvi \citep{Rossi2019} proposes a way to generate non-factorized distributions with reduced requirements in memory and computational complexity.
By considering a prior for the elements of the diagonal matrix $\Gmatr = \diag(\gvect)$ and a variational posterior $q(\gvect) = \N(\muvect, \Sigmamatr)$, we can  obtain a class of approximate posterior with some desirable properties.
%
%
%
Let $\Wmatr \in \R^{D \times D}$ be the weight matrix and consider
\begin{align}
    \tilde\Wmatr \sim q(\Wmatr) \quad \mathrm{s.t.} \quad \tilde\Wmatr = \Smatr_1 \Hmatr \diag(\tilde\gvect) \Hmatr\Smatr_2 \quad \mathrm{with} \quad \tilde\gvect \sim q(\gvect).
\end{align}

\setlength\figureheight{.3\textwidth}
\setlength\figurewidth{.3\textwidth}
\begin{minipage}[t]{.6\textwidth}
    The choice of a Gaussian $q(\gvect)$ and the linearity of the operations, induce a parameterization of a matrix-variate Gaussian distribution for $\Wmatr$, which is controlled by $\Smatr_1$ and $\Smatr_2$. These diagonal matrices can be optimized during the training.
    We refer the Reader to check \citep{Rossi2019} for an extended analysis of this factorization.
Sampling from such a distribution is achieved with the so-called {\em reparameterization trick} \citep{Kingma14}. 
As we assume a Gaussian posterior for $\gvect$, the expression $\gvect = \muvect + \Sigmamatr^{1/2}\varepsilonvect$ separates out the stochastic component ($\varepsilonvect\sim\norm(\zerovect,\Imatr)$) from the deterministic ones ($\muvect$ and $\Sigmamatr^{1/2}$).

\end{minipage}
\hfill
\begin{minipage}[t]{.375\textwidth}
    \vspace{-3ex}
    \tiny
    \pgfplotsset{every axis title/.append style={yshift=-1ex}}
    \pgfplotsset{every x tick label/.append style={font=\fontsize{2}{4}\selectfont}}
    \pgfplotsset{every y tick label/.append style={font=\fontsize{2}{4}\selectfont}}
    \sc
    \loadtikz{figures/1d_covmatrix/cov_matrix}
    \captionof{figure}{Normalized covariance of $\gvect$ and $\vect(\Wmatr)$}
    \label{fig:covariance}
\end{minipage}
To reduce the variance of stochastic gradients in the optimization and improving convergence, we also report the formulation of the {\em local reparameterization trick} \citep{Kingma2015}, which, given some input vectors $\hvect$, considers the distribution of the product $\Wmatr \hvect$.
The product $\Wmatr \hvect$ follows the distribution $\norm(\mvect,\Amatr\Amatr^\T)$ \citep{Gupta1999}, with
\begin{equation}
    \mvect = \Smatr_1 \Hmatr  \diag(\muvect) \Hmatr \Smatr_2 \hvect \text{,} \quad \text{and} \quad
    \Amatr = \Smatr_1 \Hmatr  \diag( \Hmatr \Smatr_2 \hvect)  \Sigmamatr^{1/2}.
\end{equation}
A sample from this distribution 
can be efficiently computed thanks to the Walsh-Hadamard transform as: 
$
    \overline\Wmatr(\muvect) \hvect + \overline\Wmatr(\Sigmavect^{1/2}\varepsilonvect)\hvect \text{,}
$
with $\overline\Wmatr$ a linear matrix-valued function $\overline\Wmatr(\uvect) =  \Smatr_1 \Hmatr  \diag(\uvect) \Hmatr \Smatr_2$.

\section{Experiments}
\label{sec:experiments}


\subsection{Bayesian Neural Networks}
\begin{table}[!b]
    \sc
    \scriptsize
    \centering
    \caption{Test \rmse and test \mnll for regression datasets}
    \label{tab:regression_bnn}
    \begin{tabular}{l||rrr|rrr}
    \toprule
    {}         & \multicolumn{3}{r|}{{test error}} & \multicolumn{3}{r}{{test mnll}}                                                                                               \\
    model      & {mcd}                             & nng                             & \hvi               & mcd                  & nng                   & \hvi                 \\
    dataset    &                                   &                                 &                      &                      &                       &                       \\
    \midrule
    boston     & $ 3.91 \pm 0.86$                  & $3.56 \pm 0.43$                 & $\bm{3.14 \pm 0.71}$ & $ 6.90 \pm 2.93$     & $\bm{ 2.72 \pm 0.09}$ & $ 4.33 \pm 1.80$      \\
    concrete   & $ 5.12 \pm 0.79$                  & $8.21 \pm 0.55$                 & $\bm{4.70 \pm 0.72}$ & $ 3.20 \pm 0.36$     & $ 3.56 \pm 0.08$      & $\bm{ 3.17 \pm 0.37}$ \\
    energy     & $ 2.07 \pm 0.11$                  & $1.96 \pm 0.28$                 & $\bm{0.58 \pm 0.07}$ & $ 4.15 \pm 0.15$     & $ 2.11 \pm 0.12$      & $\bm{ 2.00 \pm 0.60}$ \\
    kin8nm     & $ 0.09 \pm 0.00$                  & $\bm{0.07 \pm 0.00}$            & $0.08 \pm 0.00$      & $-0.87 \pm 0.02$     & $\bm{-1.19 \pm 0.04}$ & $\bm{-1.19 \pm 0.04}$ \\
    naval      & $ 0.30 \pm 0.30$                  & $\bm{0.00 \pm 0.00}$            & $0.01 \pm 0.00$      & $-1.00 \pm 2.27$     & $\bm{-6.52 \pm 0.09}$ & $-6.25 \pm 0.01$      \\
    powerplant & $3.97 \pm 0.14$                  & $4.23 \pm 0.09$                 & $\bm{4.00 \pm 0.12}$ & $2.74 \pm 0.05$     & $ 2.86 \pm 0.02$      & $\bm{2.71 \pm 0.03}$  \\
    protein    & $\bm{4.23 \pm 0.10}$              & $4.57 \pm 0.47$                 & $4.36 \pm 0.11$      & $\bm{2.76 \pm 0.02}$ & $ 2.95 \pm 0.12$      & $ 2.79 \pm 0.01$      \\
    yacht      & $ 1.90 \pm 0.54$                  & $5.16 \pm 1.48$                 & $\bm{0.69 \pm 0.16}$ & $ 2.95 \pm 1.27$     & $ 3.06 \pm 0.27$      & $\bm{1.80 \pm 1.01}$  \\
    \bottomrule
\end{tabular}

\end{table}
We conduct a series of comparisons with state-of-the-art \vi schemes for Bayesian \dnns: \mcd and \name{noisy-kfac} (also referred to as \name{nng}; \citep{Zhang2018}). 
\mcd draws on a formal connection between dropout and \vi with Bernoulli-like posteriors, while the more recent \name{noisy-kfac} yields a matrix-variate Gaussian distribution using noisy natural gradients. 
In \hvi, the last layer assumes a fully factorized Gaussian posterior.

Data is randomly divided into 90\%/10\% splits for training and testing.
We standardize the input features $\xvect$ while keeping the targets $\yvect$ unnormalized.
Differently from the experimental setup in \citep{Louizos2016, Zhang2018, Lobato2015}, the network has two hidden layers and 128 features with \relu activations for all the datasets and its output is parameterized as $\yvect = \fvect(\xvect) \odot \sigmavect_y + \muvect_y$.

We report the test \rmse and the average predictive test negative log-likelihood (\mnll).
A selection of results is showed in \autoref{tab:regression_bnn}. 
On the majority of the datasets, \hvi outperforms \mcd and \name{noisy-kfac}.
Empirically, these results demonstrate the value of \hvi, which offers a competitive parameterization of a matrix-variate Gaussian posterior while requiring log-linear time in $D$.

\subsection{Gaussian Processes with Random Feature Expansion}

We test \hvi for scalable \gp inference,
by focusing on \gps with random feature expansions \citep{LazaroGredilla10, Cutajar17}.
We compare \hvi with two alternatives; one is \vi of the Fourier features \gp expansion that uses less random features to match the number of parameters used in \hvi, and another is the sparse Gaussian process implementation of \name{gpflow} \citep{GPflow2017} with a number of inducing points (rounded up) to match the number of parameters used in \hvi.

We report the results 
on five datasets ($10000\leq N\leq 200000$, $5\leq D\leq 8$), generated from space-filling evaluations of well known functions in analysis of computer experiments (see e.g. \cite{simulationlib}). Dataset splitting in training and testing points is random uniform with ratio 80\%/20\%.

\begin{figure}[t!]
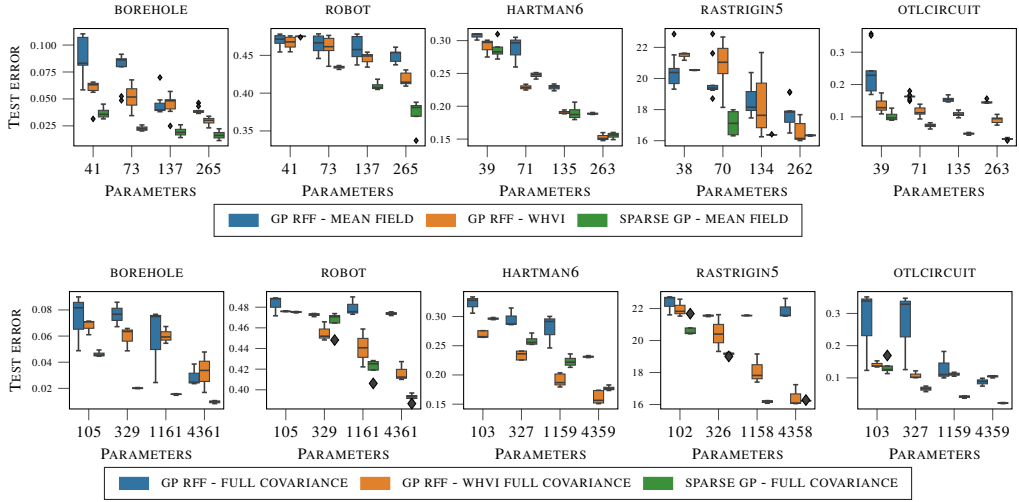

    \sc\tiny
    \centering
    \setlength\figureheight{.225\textwidth}
    \setlength\figurewidth{.26\textwidth}

    \pgfplotsset{every axis title/.append style={yshift=-1ex}}
    \pgfplotsset{every y tick label/.append style={font=\fontsize{4}{4}\selectfont}}
    \begin{subfigure}[]{\textwidth}
        \centering
        \loadtikz{figures/gp_wrt_nparameters/gp-fastfood-fulldiag.tex}
        \loadtikz{figures/gp_wrt_nparameters/legend-fulldiag.tex}
    \end{subfigure}
    \begin{subfigure}[]{\textwidth}
        \centering
        \vspace{3ex}
        \loadtikz{figures/gp_wrt_nparameters/gp-fastfood-fullcov.tex}
        \loadtikz{figures/gp_wrt_nparameters/legend-fullcov.tex}
    \end{subfigure}
    \caption{Comparison of test errors with respect to the number model parameters.}
    \label{fig:vfastgp_wrt_nparam}
\end{figure}

The results are shown in Figure~\ref{fig:vfastgp_wrt_nparam} for both with diagonal covariance and with full covariance over the model parameters.
In both mean field and full covariance settings, this variant of \hvi using the reshaping of $\Wmatr$ into a column largely outperforms the direct \vi of Fourier features.
However, it appears that this improvement of the random feature inference for \gps is still not enough to reach the performance of \vi using inducing points.
Inducing point approximations are based on the Nystr\"om approximation of kernel matrices, which are known to lead to lower approximation error on the elements on the kernel matrix compared to random features approximations.
This is the reason we attribute to the lower performance of \hvi compared to inducing points approximations in this experiment.



\section{Discussion and Conclusions}
\label{sec:conclusions}
Inspired by the literature on scalable kernel methods, this paper proposed Walsh-Hadamard Variational Inference (\hvi). 
\hvi offers a novel parameterization of the variational posterior as it assumes a matrix-variate posterior distribution, which therefore captures covariances across weights.
Crucially, unlike previous work on matrix-variate posteriors for \vi, this is achieved with a low parameterization and fast computations, bypassing the over-regularization issues of \vi for over-parameterized models. 


The key operation that contributes to accelerate computations in \hvi is the Walsh-Hadamard transform. 
This has obvious connections with other matrix/vector operations, such as the Discrete Fourier Transform and other circulant matrixes \citep{Ding2017, Cheng2015}, so we are currently investigating whether it is possible to generalize \hvi to other kinds of transforms to increase flexibility. 
%
Finally, we are looking into employing \hvi for other models, such as deep generative models.

\paragraph{Acknowledgements}
MF gratefully acknowledges support from the AXA Research Fund.

\FloatBarrier
\small

\begin{thebibliography}{33}
\providecommand{\natexlab}[1]{#1}
\providecommand{\url}[1]{\texttt{#1}}
\expandafter\ifx\csname urlstyle\endcsname\relax
  \providecommand{\doi}[1]{doi: #1}\else
  \providecommand{\doi}{doi: \begingroup \urlstyle{rm}\Url}\fi

\bibitem[Bojarski et~al.(2017)Bojarski, Choromanska, Choromanski, Fagan,
  Gouy-Pailler, Morvan, Sakr, Sarlos, and Atif]{Bojarski2017}
M.~Bojarski, A.~Choromanska, K.~Choromanski, F.~Fagan, C.~Gouy-Pailler,
  A.~Morvan, N.~Sakr, T.~Sarlos, and J.~Atif.
\newblock {Structured Adaptive and Random Spinners for Fast Machine Learning
  Computations}.
\newblock In A.~Singh and J.~Zhu, editors, \emph{Proceedings of the 20th
  International Conference on Artificial Intelligence and Statistics},
  volume~54 of \emph{Proceedings of Machine Learning Research}, pages
  1020--1029, Fort Lauderdale, FL, USA, 20--22 Apr 2017. PMLR.

\bibitem[Bowman et~al.(2016)Bowman, Vilnis, Vinyals, Dai, Jozefowicz, and
  Bengio]{Bowman2016}
S.~R. Bowman, L.~Vilnis, O.~Vinyals, A.~Dai, R.~Jozefowicz, and S.~Bengio.
\newblock {Generating Sentences from a Continuous Space}.
\newblock In \emph{Proceedings of The 20th SIGNLL Conference on Computational
  Natural Language Learning}, pages 10--21. Association for Computational
  Linguistics, 2016.

\bibitem[{Cheng} et~al.(2015){Cheng}, {Yu}, {Feris}, {Kumar}, {Choudhary}, and
  {Chang}]{Cheng2015}
Y.~{Cheng}, F.~X. {Yu}, R.~S. {Feris}, S.~{Kumar}, A.~{Choudhary}, and
  S.~{Chang}.
\newblock {An Exploration of Parameter Redundancy in Deep Networks with
  Circulant Projections}.
\newblock In \emph{2015 IEEE International Conference on Computer Vision
  (ICCV)}, pages 2857--2865, Dec 2015.

\bibitem[Cho and Saul(2009)]{Cho09}
Y.~Cho and L.~K. Saul.
\newblock {Kernel Methods for Deep Learning}.
\newblock In Y.~Bengio, D.~Schuurmans, J.~D. Lafferty, C.~K.~I. Williams, and
  A.~Culotta, editors, \emph{Advances in Neural Information Processing Systems
  22}, pages 342--350. Curran Associates, Inc., 2009.

\bibitem[Cutajar et~al.(2017)Cutajar, Bonilla, Michiardi, and
  Filippone]{Cutajar17}
K.~Cutajar, E.~V. Bonilla, P.~Michiardi, and M.~Filippone.
\newblock Random feature expansions for deep {G}aussian processes.
\newblock In D.~Precup and Y.~W. Teh, editors, \emph{Proceedings of the 34th
  International Conference on Machine Learning}, volume~70 of \emph{Proceedings
  of Machine Learning Research}, pages 884--893, International Convention
  Centre, Sydney, Australia, Aug. 2017. PMLR.

\bibitem[Damianou and Lawrence(2013)]{Damianou13}
A.~C. Damianou and N.~D. Lawrence.
\newblock {Deep Gaussian Processes}.
\newblock In \emph{Proceedings of the Sixteenth International Conference on
  Artificial Intelligence and Statistics, {AISTATS} 2013, Scottsdale, AZ, USA,
  April 29 - May 1, 2013}, volume~31 of \emph{{JMLR} Proceedings}, pages
  207--215. JMLR.org, 2013.

\bibitem[{Ding} et~al.(2017){Ding}, {Liao}, {Wang}, {Li}, {Liu}, {Zhuo},
  {Wang}, {Qian}, {Bai}, {Yuan}, {Ma}, {Zhang}, {Tang}, {Qiu}, {Lin}, and
  {Yuan}]{Ding2017}
C.~{Ding}, S.~{Liao}, Y.~{Wang}, Z.~{Li}, N.~{Liu}, Y.~{Zhuo}, C.~{Wang},
  X.~{Qian}, Y.~{Bai}, G.~{Yuan}, X.~{Ma}, Y.~{Zhang}, J.~{Tang}, Q.~{Qiu},
  X.~{Lin}, and B.~{Yuan}.
\newblock {CirCNN: Accelerating and Compressing Deep Neural Networks Using
  Block-Circulant Weight Matrices}.
\newblock In \emph{2017 50th Annual IEEE/ACM International Symposium on
  Microarchitecture (MICRO)}, pages 395--408, Oct 2017.

\bibitem[{Fino} and {Algazi}(1976)]{Fino1976}
{Fino} and {Algazi}.
\newblock {Unified Matrix Treatment of the Fast Walsh-Hadamard Transform}.
\newblock \emph{IEEE Transactions on Computers}, C-25\penalty0 (11):\penalty0
  1142--1146, Nov 1976.
\newblock ISSN 0018-9340.

\bibitem[Gal and Ghahramani(2016)]{Gal16}
Y.~Gal and Z.~Ghahramani.
\newblock {Dropout As a Bayesian Approximation: Representing Model Uncertainty
  in Deep Learning}.
\newblock In \emph{Proceedings of the 33rd International Conference on
  International Conference on Machine Learning - Volume 48}, ICML'16, pages
  1050--1059. JMLR.org, 2016.

\bibitem[Graves(2011)]{Graves11}
A.~Graves.
\newblock {Practical Variational Inference for Neural Networks}.
\newblock In J.~Shawe-Taylor, R.~S. Zemel, P.~L. Bartlett, F.~Pereira, and
  K.~Q. Weinberger, editors, \emph{Advances in Neural Information Processing
  Systems 24}, pages 2348--2356. Curran Associates, Inc., 2011.

\bibitem[Gupta and Nagar(1999)]{Gupta1999}
A.~K. Gupta and D.~K. Nagar.
\newblock \emph{Matrix variate distributions}.
\newblock Chapman and Hall/CRC, 1999.

\bibitem[Hernandez-Lobato and Adams(2015)]{Lobato2015}
J.~M. Hernandez-Lobato and R.~Adams.
\newblock Probabilistic backpropagation for scalable learning of bayesian
  neural networks.
\newblock In F.~Bach and D.~Blei, editors, \emph{Proceedings of the 32nd
  International Conference on Machine Learning}, volume~37 of \emph{Proceedings
  of Machine Learning Research}, pages 1861--1869, Lille, France, 07--09 Jul
  2015. PMLR.

\bibitem[Jordan et~al.(1999)Jordan, Ghahramani, Jaakkola, and Saul]{Jordan99}
M.~I. Jordan, Z.~Ghahramani, T.~S. Jaakkola, and L.~K. Saul.
\newblock {An Introduction to Variational Methods for Graphical Models}.
\newblock \emph{Machine Learning}, 37\penalty0 (2):\penalty0 183--233, Nov.
  1999.

\bibitem[Kingma and Welling(2014)]{Kingma14}
D.~P. Kingma and M.~Welling.
\newblock {Auto-Encoding Variational Bayes}.
\newblock In \emph{Proceedings of the Second International Conference on
  Learning Representations (ICLR 2014)}, Apr. 2014.

\bibitem[Kingma et~al.(2015)Kingma, Salimans, and Welling]{Kingma2015}
D.~P. Kingma, T.~Salimans, and M.~Welling.
\newblock {Variational Dropout and the Local Reparameterization Trick}.
\newblock In \emph{Advances in Neural Information Processing Systems 28}, pages
  2575--2583. Curran Associates, Inc., 2015.

\bibitem[L\'{a}zaro-Gredilla et~al.(2010)L\'{a}zaro-Gredilla,
  Quinonero-Candela, Rasmussen, and Figueiras-Vidal]{LazaroGredilla10}
M.~L\'{a}zaro-Gredilla, J.~Quinonero-Candela, C.~E. Rasmussen, and A.~R.
  Figueiras-Vidal.
\newblock {Sparse Spectrum Gaussian Process Regression}.
\newblock \emph{Journal of Machine Learning Research}, 11:\penalty0 1865--1881,
  2010.

\bibitem[Le et~al.(2013)Le, Sarlos, and Smola]{Le13}
Q.~Le, T.~Sarlos, and A.~Smola.
\newblock {Fastfood - Approximating Kernel Expansions in Loglinear Time}.
\newblock In \emph{30th International Conference on Machine Learning (ICML)},
  2013.

\bibitem[Louizos and Welling(2016)]{Louizos2016}
C.~Louizos and M.~Welling.
\newblock {Structured and Efficient Variational Deep Learning with Matrix
  Gaussian Posteriors}.
\newblock In M.~F. Balcan and K.~Q. Weinberger, editors, \emph{Proceedings of
  The 33rd International Conference on Machine Learning}, volume~48 of
  \emph{Proceedings of Machine Learning Research}, pages 1708--1716, New York,
  New York, USA, 20--22 Jun 2016. PMLR.

\bibitem[Louizos et~al.(2017)Louizos, Ullrich, and Welling]{Louizos2017a}
C.~Louizos, K.~Ullrich, and M.~Welling.
\newblock {Bayesian Compression for Deep Learning}.
\newblock In I.~Guyon, U.~V. Luxburg, S.~Bengio, H.~Wallach, R.~Fergus,
  S.~Vishwanathan, and R.~Garnett, editors, \emph{Advances in Neural
  Information Processing Systems 30}, pages 3288--3298. Curran Associates,
  Inc., 2017.

\bibitem[Mac{k}ay(1994)]{Mackay94}
D.~J.~C. Mac{k}ay.
\newblock {B}ayesian methods for backpropagation networks.
\newblock In E.~Domany, J.~L. van Hemmen, and K.~Schulten, editors,
  \emph{Models of Neural Networks {III}}, chapter~6, pages 211--254. Springer,
  1994.

\bibitem[Matthews et~al.(2017)Matthews, {van der Wilk}, Nickson, Fujii,
  {Boukouvalas}, {Le{\'o}n-Villagr{\'a}}, Ghahramani, and Hensman]{GPflow2017}
A.~G. d.~G. Matthews, M.~{van der Wilk}, T.~Nickson, K.~Fujii,
  A.~{Boukouvalas}, P.~{Le{\'o}n-Villagr{\'a}}, Z.~Ghahramani, and J.~Hensman.
\newblock {GP}flow: A {G}aussian process library using {T}ensor{F}low.
\newblock \emph{Journal of Machine Learning Research}, 18\penalty0
  (40):\penalty0 1--6, apr 2017.

\bibitem[Molchanov et~al.(2017)Molchanov, Ashukha, and Vetrov]{Molchanov2017}
D.~Molchanov, A.~Ashukha, and D.~Vetrov.
\newblock {Variational Dropout Sparsifies Deep Neural Networks}.
\newblock In D.~Precup and Y.~W. Teh, editors, \emph{Proceedings of the 34th
  International Conference on Machine Learning}, volume~70 of \emph{Proceedings
  of Machine Learning Research}, pages 2498--2507, International Convention
  Centre, Sydney, Australia, 06--11 Aug 2017. PMLR.

\bibitem[Neal(1996)]{Neal1997}
R.~M. Neal.
\newblock \emph{{Bayesian Learning for Neural Networks}}.
\newblock Springer-Verlag, Berlin, Heidelberg, 1996.
\newblock ISBN 0387947248.

\bibitem[Rahimi and Recht(2008)]{Rahimi08}
A.~Rahimi and B.~Recht.
\newblock {Random Features for Large-Scale Kernel Machines}.
\newblock In J.~C. Platt, D.~Koller, Y.~Singer, and S.~T. Roweis, editors,
  \emph{Advances in Neural Information Processing Systems 20}, pages
  1177--1184. Curran Associates, Inc., 2008.

\bibitem[Rezende and Mohamed(2015)]{Rezende2015}
D.~Rezende and S.~Mohamed.
\newblock {Variational Inference with Normalizing Flows}.
\newblock In F.~Bach and D.~Blei, editors, \emph{Proceedings of the 32nd
  International Conference on Machine Learning}, volume~37 of \emph{Proceedings
  of Machine Learning Research}, pages 1530--1538, Lille, France, 07--09 Jul
  2015. PMLR.

\bibitem[Rossi et~al.(2019{\natexlab{a}})Rossi, Marmin, and
  Filippone]{Rossi2019}
S.~Rossi, S.~Marmin, and M.~Filippone.
\newblock {Walsh-Hadamard Variational Inference for Bayesian Deep Learning}.
\newblock In \emph{arXiv: 1905.11248}, 2019{\natexlab{a}}.

\bibitem[Rossi et~al.(2019{\natexlab{b}})Rossi, Michiardi, and
  Filippone]{Rossi2018}
S.~Rossi, P.~Michiardi, and M.~Filippone.
\newblock {Good Initializations of Variational {B}ayes for Deep Models}.
\newblock In K.~Chaudhuri and R.~Salakhutdinov, editors, \emph{Proceedings of
  the 36th International Conference on Machine Learning}, volume~97 of
  \emph{Proceedings of Machine Learning Research}, pages 5487--5497, Long
  Beach, California, USA, 09--15 Jun 2019{\natexlab{b}}. PMLR.

\bibitem[S{\o}nderby et~al.(2016)S{\o}nderby, Raiko, Maal{\o}e, S{\o}nderby,
  and Winther]{Sonderby2016}
C.~K. S{\o}nderby, T.~Raiko, L.~Maal{\o}e, S.~K. S{\o}nderby, and O.~Winther.
\newblock {Ladder Variational Autoencoders}.
\newblock In D.~D. Lee, M.~Sugiyama, U.~V. Luxburg, I.~Guyon, and R.~Garnett,
  editors, \emph{Advances in Neural Information Processing Systems 29}, pages
  3738--3746. Curran Associates, Inc., 2016.

\bibitem[Surjanovic and Bingham()]{simulationlib}
S.~Surjanovic and D.~Bingham.
\newblock Virtual library of simulation experiments: Test functions and
  datasets.
\newblock Retrieved May 22, 2019, from \url{http://www.sfu.ca/~ssurjano}.

\bibitem[Tropp(2011)]{Tropp2011}
J.~A. Tropp.
\newblock {Improved Analysis of the subsampled Randomized Hadamard Transform}.
\newblock \emph{{Advances in Adaptive Data Analysis}}, 3\penalty0
  (1-2):\penalty0 115--126, 2011.

\bibitem[Yu et~al.(2016{\natexlab{a}})Yu, Suresh, Choromanski, Holtmann-Rice,
  and Kumar]{Yu16}
F.~X. Yu, A.~T. Suresh, K.~M. Choromanski, D.~N. Holtmann-Rice, and S.~Kumar.
\newblock {Orthogonal Random Features}.
\newblock In D.~D. Lee, M.~Sugiyama, U.~V. Luxburg, I.~Guyon, and R.~Garnett,
  editors, \emph{Advances in Neural Information Processing Systems 29}, pages
  1975--1983. Curran Associates, Inc., 2016{\natexlab{a}}.

\bibitem[Yu et~al.(2016{\natexlab{b}})Yu, Suresh, Choromanski, Holtmann-Rice,
  and Kumar]{Yu2016}
F.~X.~X. Yu, A.~T. Suresh, K.~M. Choromanski, D.~N. Holtmann-Rice, and
  S.~Kumar.
\newblock {Orthogonal Random Features}.
\newblock In D.~D. Lee, M.~Sugiyama, U.~V. Luxburg, I.~Guyon, and R.~Garnett,
  editors, \emph{Advances in Neural Information Processing Systems 29}, pages
  1975--1983. Curran Associates, Inc., 2016{\natexlab{b}}.

\bibitem[Zhang et~al.(2018)Zhang, Sun, Duvenaud, and Grosse]{Zhang2018}
G.~Zhang, S.~Sun, D.~Duvenaud, and R.~Grosse.
\newblock {Noisy Natural Gradient as Variational Inference}.
\newblock In J.~Dy and A.~Krause, editors, \emph{Proceedings of the 35th
  International Conference on Machine Learning}, volume~80 of \emph{Proceedings
  of Machine Learning Research}, pages 5852--5861, Stockholmsmässan, Stockholm
  Sweden, 10--15 Jul 2018. PMLR.

\end{thebibliography}

\end{document}